\documentclass[runningheads]{llncs}
\usepackage{graphicx}
\usepackage[font=small,labelfont=bf]{caption}
\usepackage{tabularx}
\usepackage{booktabs}
\usepackage{amsmath}
\usepackage{cleveref}
\usepackage[T1]{fontenc}
\usepackage{tikz}
\usepackage[draft]{tikzpeople}
\usepackage{acronym}
\usepackage{xspace}
\usepackage{todonotes}
\usepackage{cite}
\usepackage{array}
\usepackage{csquotes}

\usetikzlibrary{calc,positioning,shapes.geometric}
\usetikzlibrary{patterns}
\usetikzlibrary{arrows}
\usetikzlibrary{shapes}
\usetikzlibrary{fit}

\pgfdeclarelayer{bg}
\pgfsetlayers{bg,main}

\newacro{EDA}{exploratory data analysis}
\newacro{ML}{machine learning}
\newacro{PPDM}{privacy-preserving data mining}
\newacro{PPDP}{privacy-preserving data publishing}
\newacro{crosstab}{cross-tabulation table}
\newacro{GDPR}{General Data Protection Regulation}
\newacro{PPML}{privacy-preserving machine learning}

\newcommand*{\ie}{i.e.\@\xspace}

\begin{document}
\title{Every Query Counts: Analyzing the Privacy Loss of Exploratory Data 
Analyses}
\titlerunning{Privacy Loss of Exploratory Data Analyses}
\author{Saskia Nu\~nez von Voigt\inst{1} \and
Mira Pauli\inst{2} \and Johanna Reichert\inst{2} \and
Florian Tschorsch\inst{1}}
\authorrunning{S. Nu\~nez von Voigt et al.}

\institute{Distributed Security Infrastructure Group, Technische Universit\"at
Berlin,\\ Stra{\ss}e des 17. Juni 135, 10623 Berlin, Germany\\
\email{\{saskia.nunezvonvoigt,florian.tschorsch\}@tu-berlin.de}\\
\and
anacision GmbH,\\ Albert-Nestler-Stra{\ss}e 19, 76131 Karlsruhe, Germany\\
\email{\{mira.pauli,johanna.reichert\}@anacision.de}}
\maketitle              \begin{abstract}
An exploratory data analysis is an essential step for every data analyst to 
gain insights, evaluate data quality and (if required) select a machine 
learning model for further processing.
While privacy-preserving machine learning is on the rise,
more often than not this initial analysis is not counted towards the privacy 
budget.
In this paper, we quantify the privacy loss for basic statistical functions
and highlight the importance of taking it into account
when calculating the privacy-loss budget of a machine learning approach.
\end{abstract}

\section{Introduction}

One of the most prevalent barriers of machine learning
involve data management in general
and information security and privacy in particular.
This is especially relevant for sensitive data sets that,
for example, include medical and financial data items.
In order to overcome the barriers,
the area of \ac{PPML}
gained attention~\cite{Al-RubaieC19,MohasselZ17}.
It is concerned with providing an
infrastructure for secure and privacy-preserving data access
as well as privacy-preserving model generation.

While \ac{PPML} reduces the risk of data leaks,
particularly the risk of model inversion attacks,
one aspect is often overlooked:
In order to decide which type of model should be trained
and how it should be parametrized,
a data analyst performs a preceding \ac{EDA}.
The \ac{EDA} consists of querying the data for a number of statistics and 
metrics.
The goal is to gain insights on the  data quality, as well as the relationships 
between the  variables
to initiate data preprocessing
\emph{before} a model is created.
To do so, the data analyst has typically full data access
and is not restricted in her queries.

In this paper, we evaluate the privacy loss of performing an \ac{EDA}.
To this end, we assume that an analyst obtains differentially-private 
answers~\cite{Dwork06}
to preserve information privacy.
We quantify the privacy loss for basic analysis steps,
which can be used to make decisions on how to clean the data,
select features, and to select a model.
Based on our evaluation,
we discuss the implications of the resulting privacy loss and conclude that
an interactive \ac{EDA} is not feasible in a privacy-preserving setting.

At the same time, we compare the accuracy of an interactive approach
with the generation of differentially-private synthetic data.
Our results underline that the privacy loss can be mitigated by determining
which functions are needed such that they can be answered as correlated queries.
The generation of synthetic data is a generalization of this approach.
Accordingly, we recommend to develop standardized sets of \ac{EDA} functions
to reduce the privacy loss and/or increase accuracy.
In all cases, however, the privacy loss inevitably increases
with the amount of information requested and
should be considered for \acp{EDA} in general.

This paper is organized as follows. In \Cref{sec:systemmodell} and 
\Cref{sec:eda},
we describe our system model
and introduce a basic set of \ac{EDA} functions, respectively.
In \Cref{sec:eval}, we evaluate the privacy loss and discuss the
feasibility of a differentially-private interactive \ac{EDA},
before concluding this paper in \Cref{sec:conclusion}.

\section{Background}
\label{sec:systemmodell}

\subsection{System and Adversary Model}
\begin{figure}[t]
	\centering

\begin{tikzpicture}[node distance=0pt and 35mm]

	    \tikzstyle{every node}=[font=\scriptsize]

		\node[] (db) {\includegraphics[height=12mm]{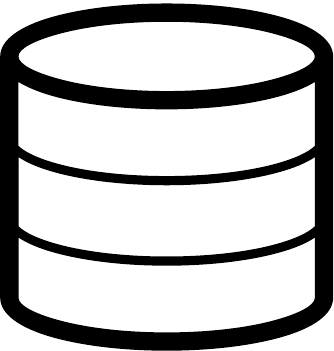}};
  		  \node[below=of db,align=center] {Data set};
\node[fit=(db.north east)(db.south east),inner sep=5pt,draw,
  		  thick,xshift=40pt]  (dp) {};
  		\node[rotate=90] at(dp.base) {privacy};
  		  \node[above=of dp,align=center,yshift=15pt,text width=25mm]
  		  (budget) {privacy budget $\epsilon$};

\node[right=of db,xshift=10pt] (analyst)
		{\includegraphics[height=18mm]{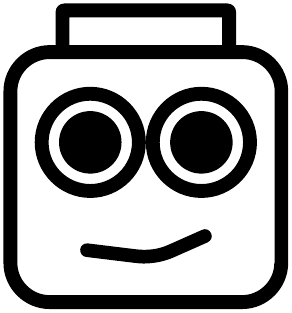}};
  		  \node[below=of analyst,align=center] {Data analyst\\(adversary)};
		\node[right=of analyst] (model)
		{\includegraphics[height=12mm]{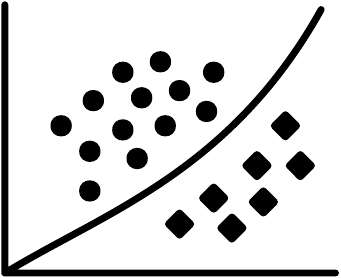}};
  		  \node[below=of model,align=center] {Model};

		\draw[->,thick] (dp.north) -- node[right,scale=0.8,pos=.5]
		  {$-\epsilon_i$}(budget);
  		\draw[->,thick,transform canvas={yshift=6pt},shorten >=5pt]
  		  (analyst) -- node[above,scale=0.8] {$\text{query}_1$,$\epsilon_1=1$}
  		  (dp);
  		\draw[->,thick,transform canvas={yshift=6pt},shorten <=5pt]
  		  (dp) -- node[above,scale=0.8] {$\text{query}_1$}
  		  (db);
  		\draw[->,thick,transform canvas={yshift=-6pt},shorten >=5pt]
  		  (db) -- node[above,scale=0.8] {$\text{result}_1$}
  		  (dp);
  		\draw[->,thick,transform canvas={yshift=-6pt},shorten <=5pt]
  		  (dp) -- node[above,scale=0.8] {$\text{result}_1 + \text{noise}$} 
  		  (analyst);
  		\draw[->,thick,transform canvas={yshift=-6pt}]
  		  (analyst) -- node[above,scale=0.8,text width=26mm]
  		  {analyst performs an EDA before training a model with $\epsilon_2$}
  		  (model);
	\end{tikzpicture}
	\caption{System model.}
	\label{fig:systemmodel}
\end{figure}
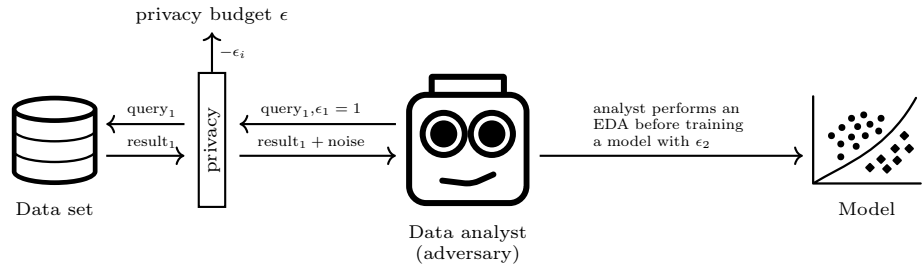

An \ac{EDA} is typically performed before a machine learning model is created
in order to obtain a basic understanding of variable distributions, data 
quality,
and the relationship between variables.
By querying this information, a data analyst can determine the
necessary steps for data cleaning and a suitable model.

In this paper, we consider the data analyst as an adversary,
who should not be able to reveal information about individuals.
That is, we assume an interactive query-response setting,
which is visualized in \Cref{fig:systemmodel}.
The analyst (internal or external)
is allowed to query a data set and request aggregated data to perform the
\ac{EDA} and afterwards train a model.
In order to mitigate re-identification attacks,
noise is added to the results, which satisfy the definition of differential 
privacy
(see below).
A privacy budget tracks the privacy loss generated by the queries. It decreases
with each query until it is spent and no further
queries are answered.
The system model captures the privacy-utility trade-off
inherent to the notion of information privacy.
We believe that the requirement of an \ac{EDA} is often overlooked
when it comes to creating privacy-preserving models.

\subsection{Differential Privacy}
\label{sec:dp}
Differential privacy quantifies the privacy loss~\cite{Dwork06}
regardless of an adversary's knowledge.
It determines the risk of being identified in a database by
comparing results of querying the database with and without
the individual concerned.
The intuition is that the absence/existence of a data subject
should have a small impact on the results,
which in turn implies that an adversary cannot identify individuals from the 
result.
More formally, a mechanism $\mathcal{K}$ provides $\epsilon$-differential 
privacy
if for all data sets $D_1$ and $D_2$, differing on at most one data subject,
and all
$S\subseteq \text{Range}(\mathcal{K})$ satisfy
\begin{align}
P[\mathcal{K}(D_1) \in S] \leq \mathrm{e}^{\epsilon} \cdot P[\mathcal{K}(D_2) 
\in S]
\text{.}
\label{eq:dp}
\end{align}
Differential privacy ensures that the result of an analysis changes
by at most a multiplicative factor $\mathrm{e}^{\epsilon}$ when a record is
included in the data set or not.  For $\epsilon=0$, the result of an analysis is exactly the same
whether a record is included or not and thus provides perfect
privacy. With $\epsilon=0$, however, we cannot obtain meaningful
results. In contrast, higher $\epsilon$ provide lower privacy guarantee.
It is therefore necessary to find a balance between
$\epsilon$ and the accuracy of the results.

Any mechanism guaranteeing differential privacy is robust under
composition~\cite{McSherryT07}. If we apply the mechanisms $\mathcal{K}_i$, each providing
$\epsilon_i$-differential privacy, several times to the same data, the sequence
of queries gives $\epsilon$-differential privacy with
$\epsilon=\sum_{i}\epsilon_i$. In other words, the maximum privacy loss is
bounded by the privacy budget
$\epsilon$, which in turn is reduced by $\epsilon_i$ for each query. 
As soon as the budget is spent, no further queries are answered.
The parallel application of mechanisms $\mathcal{K}_i$ for $D_i$, an
arbitrary disjoint subset of the input domain D, each providing
$\epsilon_i$-differential privacy, gives $\epsilon$~-differential privacy with
$\epsilon=\max(\epsilon_i)$.

In an \ac{EDA}, a data analyst queries interactively,
\ie, we assume random queries that are unknown in advance.
Therefore, we apply the differential privacy
mechanism for each query and calculate the required privacy budget
$\epsilon$ according the composition theorems.

To satisfy differential privacy for numeric queries, commonly random noise drawn
from a Laplace distribution is added to the numerical output $f(X)$~\cite{Dwork06}. 
The magnitude of noise is calibrated according to the
sensitivity of a function. The sensitivity $\Delta f$ is the maximum difference
that an output can change by removing or adding a record.
For example, a simple counting query, \ie, how many rows have a specific 
variable value, has
$\Delta f =1$.
Differential privacy is then provided by $f(X)+\text{Lap}(\Delta 
f\mathbin{/}\epsilon)$.

\section{Exploratory Data Analysis}
\label{sec:eda}
An \ac{EDA} is an essential step in any data science application.
Generally, an \ac{EDA} includes different methods and can be an
exhaustive analysis within itself.
Moreover, the process is not standardized
and depends on the objective of the analysis.

In the following, we select some basic statistical functions
that serve as the \enquote{least common denominator},
which we derived from literature as well as our own practice in the field.
The selected statistical functions for this basic \ac{EDA}
are listed in \Cref{tab:EDA_steps}.
Even if we limit our \ac{EDA} to certain functions, we still assume that the
queries performed by the analyst are not
known in advance and are sent interactively depending on the results.
Since some analyses depend on the data type, we differentiate functions
between categorical and numerical variables.

\begin{table}[t]
	\centering
	\caption{Privacy loss for statistical functions of a basic \ac{EDA}.}
	\vspace{-1ex}
	\label{tab:EDA_steps}\label{tab:fct_priv_budget}
	\resizebox{\textwidth}{!}{
		\begin{tabularx}{1.15\textwidth}{XXXX}
			\toprule
			& \multicolumn{2}{c}{Statistical function} & \\
			\cmidrule(r){2-3}\\[-2ex]
			Information & Numerical (\texttt{NUM}) & Categorical (\texttt{CAT})
			&
			Privacy loss \\
			\midrule
			distribution   (\texttt{DIST}) & range, $Q_1, Q_2, Q_3$ & value
			counts & $\epsilon_i\cdot (5 \cdot n+ c)$\\
			missing values (\texttt{MISS}) & count & value counts & $\epsilon_i
			\cdot n$ \\
			outliers       (\texttt{OUTL}) & count outside cut-off	& value
			counts & $\epsilon_i \cdot n$\\
			correlation    (\texttt{CORR}) & Spearman's correlation & Cramer's
			V & $\epsilon_i \cdot
			\left({n\choose 2}+{c\choose 2}\right)$\\
			\bottomrule
	\end{tabularx}}
\end{table}

\paragraph{Distribution.}
The distribution of the data is important to understand the data.
For numerical variables, the range and quantiles provide
information about the validity of the data and a sense of the range of the
data.
For categorical variables, a data analyst
retrieves the unique variable values, especially the number of observations of 
each variable value.
Variables with a discrete uniform distribution, for example, are not suitable 
to identify
meaningful patterns. Variables that show this behavior need to be cleaned or
removed for the training process.

\paragraph{Missing Values.}
The number of missing values indicates whether steps for data cleaning are
necessary.
There are different options such as case deletions or imputation with a vast
body of literature discussing these options and their implications for later
analysis steps~\cite{acuna2004treatment}.

\paragraph{Outliers.}
Machine learning models can be influenced by outliers, thus an analyst should 
be aware
of their presence in the data set.
There are many sophisticated tools to detect outliers,
that mostly come with a high degree of
privacy loss.
Therefore, we resort to a simple box plot approach, where the cutoff point for
outliers is defined as the upper and lower quantiles ($Q_1$ and
$Q_3$) and a tolerance  of $1.5 \cdot (Q_3 - Q_1)$~\cite{laurikkala2000informal}.
We then count all values that lie above or below that cutoff point.
In this univariate outlier detection context, categorical variables are not 
covered, as rare values have already become visible from the distribution.

\paragraph{Correlation.}
The results of the correlation between
variables mainly contribute to feature selection, where certain
variables may be excluded from the model or combined with each other.
Correlated variables are problematic for the interpretation of
a model~\cite{zuur2010protocol}.
Furthermore, the relationship between independent variables may imply that
dimensionality reduction methods can be applied to the data set to improve
model performance.
Based on this motivation, we include
Spearman's correlation matrix for numerical data
and a Cramer's V for categorical variables
in our basic set of \ac{EDA} functions.

\section{Privacy loss and Accuracy Impact Assessment}
\label{sec:eval}

In \Cref{tab:fct_priv_budget}, we quantify the privacy loss
of the functions of our basic \ac{EDA}.
The privacy budget required to compute the respective functions is cumulated by
the privacy loss of each query.
It depends on the privacy loss per query $\epsilon_i$,
the number of categorical variables $c$, and
the number of numerical variables $n$.

For the numerical distribution a data analyst queries the $\min$,
$\max$, $Q_1$, $Q_2$, and $Q_3$.
In other words, to obtain the information an analyst needs to query the data
five times.
Since each record is contained in each variable, an analyst spends
$5 \cdot \epsilon_i\cdot n$ of its privacy budget for this statistical function.
The categorical distribution can be investigated by the counts per variable
value
that are queried only once per variable and the required
budget increases by $\epsilon_i$ and $c$.

The missing values as well as the outliers of categorical variables are visible
from the value
counts. Therefore, the privacy budget only increases by the numerical variables
for both the missing values and the outliers.

We quantify the relationship between two variables using Spearman's
correlation for numerical and Cramer's V for categorical variables.
Since both measures are symmetrical, the privacy budget increases by
the number of permutations

\subsection{Privacy Loss}
We determine the privacy budget for some
data sets from the UCI Machine Learning
Repository\footnote{\url{http://archive.ics.uci.edu/ml}}.
The data sets differ in size and number of variables.
Common values for $\epsilon$ comprise $0.01$, $0.1$, $1$, $\ln(2)$, or 
$\ln(3)$~\cite{Dwork08}. Therefore we fixed the privacy guarantee for each
query to $\epsilon_i=0.01$, as this is the smallest value.

\begin{figure}[t]
	\centering
	\begin{tikzpicture}[x=1pt,y=1pt]
\definecolor{fillColor}{RGB}{255,255,255}
\path[use as bounding box,fill=fillColor,fill opacity=0.00] (0,0) rectangle (346.90,173.45);
\begin{scope}
\path[clip] (  0.00,  0.00) rectangle (346.90,173.45);
\definecolor{drawColor}{RGB}{255,255,255}
\definecolor{fillColor}{RGB}{255,255,255}

\path[draw=drawColor,line width= 0.6pt,line join=round,line cap=round,fill=fillColor] (  0.00, -0.00) rectangle (346.90,173.45);
\end{scope}
\begin{scope}
\path[clip] ( 57.72, 63.98) rectangle (226.35,167.95);
\definecolor{fillColor}{RGB}{255,255,255}

\path[fill=fillColor] ( 57.72, 63.98) rectangle (226.35,167.95);
\definecolor{drawColor}{RGB}{51,34,136}

\path[draw=drawColor,line width= 0.6pt,line join=round] ( 74.04, 86.29) --
	(101.24, 94.41) --
	(128.44, 98.84) --
	(155.63,102.72) --
	(182.83,119.25) --
	(210.03,124.18);
\definecolor{drawColor}{RGB}{136,204,238}

\path[draw=drawColor,line width= 0.6pt,dash pattern=on 2pt off 2pt ,line join=round] (101.24, 68.70) --
	(182.83,136.73);
\definecolor{drawColor}{RGB}{17,119,51}

\path[draw=drawColor,line width= 0.6pt,dash pattern=on 4pt off 2pt ,line join=round] ( 74.04, 95.20) --
	(101.24,109.26) --
	(128.44,112.96) --
	(155.63,116.27) --
	(182.83,160.72) --
	(210.03,163.22);
\definecolor{drawColor}{RGB}{221,204,119}

\path[draw=drawColor,line width= 0.6pt,dash pattern=on 4pt off 4pt ,line join=round] ( 74.04,102.10) --
	(101.24,102.72) --
	(128.44,108.26) --
	(155.63,112.96) --
	(210.03,128.16);
\definecolor{drawColor}{RGB}{204,102,119}

\path[draw=drawColor,line width= 0.6pt,dash pattern=on 1pt off 3pt ,line join=round] (101.24, 78.06) --
	(182.83,155.00);
\definecolor{drawColor}{RGB}{0,0,0}

\path[draw=drawColor,line width= 1.1pt,line join=round] ( 57.72,157.58) -- (226.35,157.58);
\definecolor{drawColor}{RGB}{51,34,136}
\definecolor{fillColor}{RGB}{51,34,136}

\path[draw=drawColor,line width= 0.4pt,line join=round,line cap=round,fill=fillColor] ( 74.04, 86.29) circle (  1.43);

\path[draw=drawColor,line width= 0.4pt,line join=round,line cap=round,fill=fillColor] (101.24, 94.41) circle (  1.43);

\path[draw=drawColor,line width= 0.4pt,line join=round,line cap=round,fill=fillColor] (128.44, 98.84) circle (  1.43);

\path[draw=drawColor,line width= 0.4pt,line join=round,line cap=round,fill=fillColor] (155.63,102.72) circle (  1.43);

\path[draw=drawColor,line width= 0.4pt,line join=round,line cap=round,fill=fillColor] (182.83,119.25) circle (  1.43);

\path[draw=drawColor,line width= 0.4pt,line join=round,line cap=round,fill=fillColor] (210.03,124.18) circle (  1.43);
\definecolor{drawColor}{RGB}{136,204,238}
\definecolor{fillColor}{RGB}{136,204,238}

\path[draw=drawColor,line width= 0.4pt,line join=round,line cap=round,fill=fillColor] (101.24, 68.70) circle (  1.43);

\path[draw=drawColor,line width= 0.4pt,line join=round,line cap=round,fill=fillColor] (182.83,136.73) circle (  1.43);
\definecolor{drawColor}{RGB}{17,119,51}
\definecolor{fillColor}{RGB}{17,119,51}

\path[draw=drawColor,line width= 0.4pt,line join=round,line cap=round,fill=fillColor] ( 74.04, 95.20) circle (  1.43);

\path[draw=drawColor,line width= 0.4pt,line join=round,line cap=round,fill=fillColor] (101.24,109.26) circle (  1.43);

\path[draw=drawColor,line width= 0.4pt,line join=round,line cap=round,fill=fillColor] (128.44,112.96) circle (  1.43);

\path[draw=drawColor,line width= 0.4pt,line join=round,line cap=round,fill=fillColor] (155.63,116.27) circle (  1.43);

\path[draw=drawColor,line width= 0.4pt,line join=round,line cap=round,fill=fillColor] (182.83,160.72) circle (  1.43);

\path[draw=drawColor,line width= 0.4pt,line join=round,line cap=round,fill=fillColor] (210.03,163.22) circle (  1.43);
\definecolor{drawColor}{RGB}{221,204,119}
\definecolor{fillColor}{RGB}{221,204,119}

\path[draw=drawColor,line width= 0.4pt,line join=round,line cap=round,fill=fillColor] ( 74.04,102.10) circle (  1.43);

\path[draw=drawColor,line width= 0.4pt,line join=round,line cap=round,fill=fillColor] (101.24,102.72) circle (  1.43);

\path[draw=drawColor,line width= 0.4pt,line join=round,line cap=round,fill=fillColor] (128.44,108.26) circle (  1.43);

\path[draw=drawColor,line width= 0.4pt,line join=round,line cap=round,fill=fillColor] (155.63,112.96) circle (  1.43);

\path[draw=drawColor,line width= 0.4pt,line join=round,line cap=round,fill=fillColor] (210.03,128.16) circle (  1.43);
\definecolor{drawColor}{RGB}{204,102,119}
\definecolor{fillColor}{RGB}{204,102,119}

\path[draw=drawColor,line width= 0.4pt,line join=round,line cap=round,fill=fillColor] (101.24, 78.06) circle (  1.43);

\path[draw=drawColor,line width= 0.4pt,line join=round,line cap=round,fill=fillColor] (182.83,155.00) circle (  1.43);
\end{scope}
\begin{scope}
\path[clip] (  0.00,  0.00) rectangle (346.90,173.45);
\definecolor{drawColor}{RGB}{0,0,0}

\path[draw=drawColor,line width= 0.6pt,line join=round] ( 57.72, 63.98) --
	( 57.72,167.95);
\end{scope}
\begin{scope}
\path[clip] (  0.00,  0.00) rectangle (346.90,173.45);
\definecolor{drawColor}{gray}{0.30}

\node[text=drawColor,anchor=base east,inner sep=0pt, outer sep=0pt, scale=  0.88] at ( 52.77, 83.26) {0.3};

\node[text=drawColor,anchor=base east,inner sep=0pt, outer sep=0pt, scale=  0.88] at ( 52.77,120.53) {1.0};

\node[text=drawColor,anchor=base east,inner sep=0pt, outer sep=0pt, scale=  0.88] at ( 52.77,154.55) {3.0};
\end{scope}
\begin{scope}
\path[clip] (  0.00,  0.00) rectangle (346.90,173.45);
\definecolor{drawColor}{gray}{0.20}

\path[draw=drawColor,line width= 0.6pt,line join=round] ( 54.97, 86.29) --
	( 57.72, 86.29);

\path[draw=drawColor,line width= 0.6pt,line join=round] ( 54.97,123.56) --
	( 57.72,123.56);

\path[draw=drawColor,line width= 0.6pt,line join=round] ( 54.97,157.58) --
	( 57.72,157.58);
\end{scope}
\begin{scope}
\path[clip] (  0.00,  0.00) rectangle (346.90,173.45);
\definecolor{drawColor}{RGB}{0,0,0}

\path[draw=drawColor,line width= 0.6pt,line join=round] ( 57.72, 63.98) --
	(226.35, 63.98);
\end{scope}
\begin{scope}
\path[clip] (  0.00,  0.00) rectangle (346.90,173.45);
\definecolor{drawColor}{gray}{0.20}

\path[draw=drawColor,line width= 0.6pt,line join=round] ( 74.04, 61.23) --
	( 74.04, 63.98);

\path[draw=drawColor,line width= 0.6pt,line join=round] (101.24, 61.23) --
	(101.24, 63.98);

\path[draw=drawColor,line width= 0.6pt,line join=round] (128.44, 61.23) --
	(128.44, 63.98);

\path[draw=drawColor,line width= 0.6pt,line join=round] (155.63, 61.23) --
	(155.63, 63.98);

\path[draw=drawColor,line width= 0.6pt,line join=round] (182.83, 61.23) --
	(182.83, 63.98);

\path[draw=drawColor,line width= 0.6pt,line join=round] (210.03, 61.23) --
	(210.03, 63.98);
\end{scope}
\begin{scope}
\path[clip] (  0.00,  0.00) rectangle (346.90,173.45);
\definecolor{drawColor}{gray}{0.30}

\node[text=drawColor,rotate= 45.00,anchor=base east,inner sep=0pt, outer 
sep=0pt, scale=  0.88] at ( 78.33, 54.74) {\texttt{DIST-NUM}};

\node[text=drawColor,rotate= 45.00,anchor=base east,inner sep=0pt, outer 
sep=0pt, scale=  0.88] at (105.52, 54.74) {\texttt{DIST-CAT}};

\node[text=drawColor,rotate= 45.00,anchor=base east,inner sep=0pt, outer 
sep=0pt, scale=  0.88] at (132.72, 54.74) {\texttt{MISS-NUM}};

\node[text=drawColor,rotate= 45.00,anchor=base east,inner sep=0pt, outer 
sep=0pt, scale=  0.88] at (159.92, 54.74) {\texttt{OUTL-NUM}};

\node[text=drawColor,rotate= 45.00,anchor=base east,inner sep=0pt, outer 
sep=0pt, scale=  0.88] at (187.12, 54.74) {\texttt{CORR-CAT}};

\node[text=drawColor,rotate= 45.00,anchor=base east,inner sep=0pt, outer 
sep=0pt, scale=  0.88] at (214.32, 54.74) {\texttt{CORR-NUM}};
\end{scope}
\begin{scope}
\path[clip] (  0.00,  0.00) rectangle (346.90,173.45);
\definecolor{drawColor}{RGB}{0,0,0}

\node[text=drawColor,anchor=base,inner sep=0pt, outer sep=0pt, scale=  1] at 
(142.04,  7.44) {EDA function};
\end{scope}
\begin{scope}
\path[clip] (  0.00,  0.00) rectangle (346.90,173.45);
\definecolor{drawColor}{RGB}{0,0,0}

\node[text=drawColor,rotate= 90.00,anchor=base,inner sep=0pt, outer sep=0pt, 
scale=  1] at ( 13.08,115.96) {};

\node[text=drawColor,rotate= 90.00,anchor=base,inner sep=0pt, outer sep=0pt, 
scale=  1] at ( 24.96,115.96) {cumulative privacy loss};

\node[text=drawColor,rotate= 90.00,anchor=base,inner sep=0pt, outer sep=0pt, 
scale=  1] at ( 36.84,115.96) {};
\end{scope}
\begin{scope}
\path[clip] (  0.00,  0.00) rectangle (346.90,173.45);
\definecolor{drawColor}{RGB}{51,34,136}

\path[draw=drawColor,line width= 0.6pt,line join=round] (244.66,144.59) -- (259.11,144.59);
\end{scope}
\begin{scope}
\path[clip] (  0.00,  0.00) rectangle (346.90,173.45);
\definecolor{drawColor}{RGB}{51,34,136}
\definecolor{fillColor}{RGB}{51,34,136}

\path[draw=drawColor,line width= 0.4pt,line join=round,line cap=round,fill=fillColor] (251.88,144.59) circle (  1.43);
\end{scope}
\begin{scope}
\path[clip] (  0.00,  0.00) rectangle (346.90,173.45);
\definecolor{drawColor}{RGB}{136,204,238}

\path[draw=drawColor,line width= 0.6pt,dash pattern=on 2pt off 2pt ,line join=round] (244.66,126.52) -- (259.11,126.52);
\end{scope}
\begin{scope}
\path[clip] (  0.00,  0.00) rectangle (346.90,173.45);
\definecolor{drawColor}{RGB}{136,204,238}
\definecolor{fillColor}{RGB}{136,204,238}

\path[draw=drawColor,line width= 0.4pt,line join=round,line cap=round,fill=fillColor] (251.88,126.52) circle (  1.43);
\end{scope}
\begin{scope}
\path[clip] (  0.00,  0.00) rectangle (346.90,173.45);
\definecolor{drawColor}{RGB}{17,119,51}

\path[draw=drawColor,line width= 0.6pt,dash pattern=on 4pt off 2pt ,line join=round] (244.66,108.45) -- (259.11,108.45);
\end{scope}
\begin{scope}
\path[clip] (  0.00,  0.00) rectangle (346.90,173.45);
\definecolor{drawColor}{RGB}{17,119,51}
\definecolor{fillColor}{RGB}{17,119,51}

\path[draw=drawColor,line width= 0.4pt,line join=round,line cap=round,fill=fillColor] (251.88,108.45) circle (  1.43);
\end{scope}
\begin{scope}
\path[clip] (  0.00,  0.00) rectangle (346.90,173.45);
\definecolor{drawColor}{RGB}{221,204,119}

\path[draw=drawColor,line width= 0.6pt,dash pattern=on 4pt off 4pt ,line join=round] (244.66, 90.39) -- (259.11, 90.39);
\end{scope}
\begin{scope}
\path[clip] (  0.00,  0.00) rectangle (346.90,173.45);
\definecolor{drawColor}{RGB}{221,204,119}
\definecolor{fillColor}{RGB}{221,204,119}

\path[draw=drawColor,line width= 0.4pt,line join=round,line cap=round,fill=fillColor] (251.88, 90.39) circle (  1.43);
\end{scope}
\begin{scope}
\path[clip] (  0.00,  0.00) rectangle (346.90,173.45);
\definecolor{drawColor}{RGB}{204,102,119}

\path[draw=drawColor,line width= 0.6pt,dash pattern=on 1pt off 3pt ,line join=round] (244.66, 72.32) -- (259.11, 72.32);
\end{scope}
\begin{scope}
\path[clip] (  0.00,  0.00) rectangle (346.90,173.45);
\definecolor{drawColor}{RGB}{204,102,119}
\definecolor{fillColor}{RGB}{204,102,119}

\path[draw=drawColor,line width= 0.4pt,line join=round,line cap=round,fill=fillColor] (251.88, 72.32) circle (  1.43);
\end{scope}
\begin{scope}
\path[clip] (  0.00,  0.00) rectangle (346.90,173.45);
\definecolor{drawColor}{RGB}{0,0,0}

\node[text=drawColor,anchor=base west,inner sep=0pt, outer sep=0pt, scale=  1] 
at (266.42,141.56) {\texttt{adult}};
\end{scope}
\begin{scope}
\path[clip] (  0.00,  0.00) rectangle (346.90,173.45);
\definecolor{drawColor}{RGB}{0,0,0}

\node[text=drawColor,anchor=base west,inner sep=0pt, outer sep=0pt, scale=  1] 
at (266.42,123.49) {\texttt{congvoting}};
\end{scope}
\begin{scope}
\path[clip] (  0.00,  0.00) rectangle (346.90,173.45);
\definecolor{drawColor}{RGB}{0,0,0}

\node[text=drawColor,anchor=base west,inner sep=0pt, outer sep=0pt, scale=  1] 
at (266.42,105.42) {\texttt{drug consumption}};
\end{scope}
\begin{scope}
\path[clip] (  0.00,  0.00) rectangle (346.90,173.45);
\definecolor{drawColor}{RGB}{0,0,0}

\node[text=drawColor,anchor=base west,inner sep=0pt, outer sep=0pt, scale=  1] 
at (266.42, 87.36) {\texttt{magic04}};
\end{scope}
\begin{scope}
\path[clip] (  0.00,  0.00) rectangle (346.90,173.45);
\definecolor{drawColor}{RGB}{0,0,0}

\node[text=drawColor,anchor=base west,inner sep=0pt, outer sep=0pt, scale=  1] 
at (266.42, 69.29) {\texttt{mushroom}};
\end{scope}
\end{tikzpicture}
 	\vspace{-1em}
	\caption{Cumulative privacy loss with $\epsilon_i=0.01$ for different 
	datasets.} 
\label{fig:priv_budget}
	\vspace{-1ex}
\end{figure}
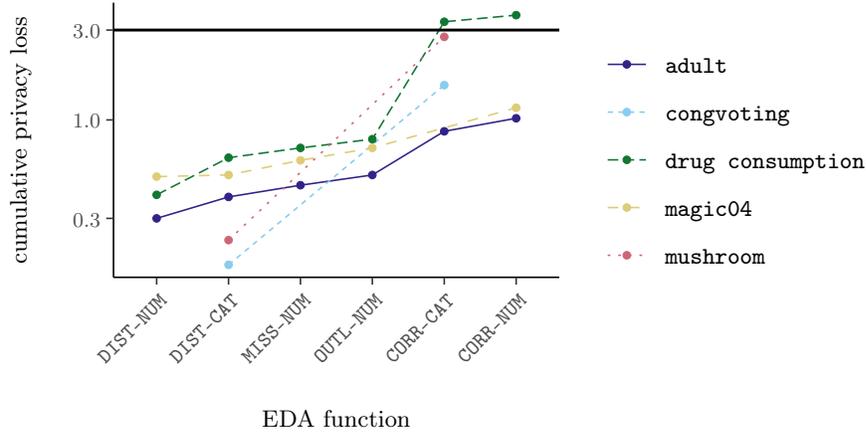

\Cref{fig:priv_budget} shows the cumulative privacy loss by
conducting
all statistical functions from our basic \ac{EDA}.
We observe an increasing and high privacy loss.
For computing all functions, the \texttt{adult} data set requires the smallest
privacy budget. The prime cause of
this difference is the small number of variables in total.
Indeed, the \texttt{magic04} data set has less variables in total
but more numerical variables.
Since an analyst sends an additional query for numerical variables to determine
the missing values or outliers, the privacy loss increases and is thus higher
than for categorical variables.

The correlation leads to the highest privacy loss, since the budget increases
by the binomial coefficient $n \choose 2$ and
$c \choose 2$.

Note that the privacy budget increases linearly with the number of queries.
With a
lower privacy guarantee, \ie, $\epsilon_i >0.01$, the total privacy budget
would exceed the privacy budget of $\epsilon = 3$, which yields a 20 times higher
chance ($\mathrm{e}^3$) to be compromised.
With a smaller $\epsilon_i$, we can reduce the privacy loss and
therefore the total required privacy budget. However, this leads to an accuracy
loss.

\subsection{Accuracy}
Due to the order of queried information the answers of our
queries cannot be re-used. In order to reduce the privacy budget,
numerous approaches
aggregate queries and determine correlations between
queries~\cite{Leoni12,LiHRMM10,LiM12,RothR10,YaroslavtsevCPS13}.
This allows estimating results from other noisy answers without spending its
privacy budget.
The results of these mechanisms can also be treated as differentially-private
synthetic data that support the original working method of a data analyst.

In this section, we evaluate the accuracy of a differentially-private \ac{EDA}
and compare the interactive setting with
differentially-private synthetic data
sets. For data sets with numeric variables, we remove some values from a
variable to have a numerical variable with $10$\% missing values.

We generate synthetic data sets using the correlated mode of
DataSynthesizer\footnote{Available for download at
\url{https://github.com/DataResponsibly/DataSynthesizer}} that learns a Bayesian
network with a degree of four.
For comparison, we generate synthetic data with the same privacy budget that is
required to investigate the correlation in an interactive setting.
For example for the \texttt{adult} synthetic data set we set
$\epsilon= 51\cdot0.01=0.51$.

\begin{figure}[t]
	\centering
	\begin{tikzpicture}[x=1pt,y=1pt]
\definecolor{fillColor}{RGB}{255,255,255}
\path[use as bounding box,fill=fillColor,fill opacity=0.00] (0,0) rectangle (346.90,289.08);
\begin{scope}
\path[clip] (  0.00,  0.00) rectangle (346.90,289.08);
\definecolor{drawColor}{RGB}{255,255,255}
\definecolor{fillColor}{RGB}{255,255,255}

\path[draw=drawColor,line width= 0.6pt,line join=round,line cap=round,fill=fillColor] (  0.00, -0.00) rectangle (346.90,289.08);
\end{scope}
\begin{scope}
\path[clip] ( 58.21,196.56) rectangle (197.05,266.78);
\definecolor{fillColor}{RGB}{255,255,255}

\path[fill=fillColor] ( 58.21,196.56) rectangle (197.05,266.78);
\definecolor{drawColor}{gray}{0.20}

\path[draw=drawColor,line width= 0.6pt,line join=round] ( 67.45,231.49) -- ( 67.45,243.18);

\path[draw=drawColor,line width= 0.6pt,line join=round] ( 67.45,210.76) -- ( 67.45,201.36);
\definecolor{fillColor}{RGB}{68,119,170}

\path[draw=drawColor,line width= 0.6pt,line join=round,line cap=round,fill=fillColor,fill opacity=0.80] ( 63.67,231.49) --
	( 63.67,210.76) --
	( 71.23,210.76) --
	( 71.23,231.49) --
	( 63.67,231.49) --
	cycle;

\path[draw=drawColor,line width= 1.1pt,line join=round] ( 63.67,217.22) -- ( 71.23,217.22);

\path[draw=drawColor,line width= 0.6pt,line join=round] ( 75.85,220.61) -- ( 75.85,223.83);

\path[draw=drawColor,line width= 0.6pt,line join=round] ( 75.85,211.91) -- ( 75.85,203.66);
\definecolor{fillColor}{RGB}{204,102,119}

\path[draw=drawColor,line width= 0.6pt,line join=round,line cap=round,fill=fillColor,fill opacity=0.80] ( 72.07,220.61) --
	( 72.07,211.91) --
	( 79.62,211.91) --
	( 79.62,220.61) --
	( 72.07,220.61) --
	cycle;

\path[draw=drawColor,line width= 1.1pt,line join=round] ( 72.07,217.53) -- ( 79.62,217.53);
\definecolor{drawColor}{RGB}{51,51,51}
\definecolor{fillColor}{RGB}{51,51,51}

\path[draw=drawColor,draw opacity=0.80,line width= 0.4pt,line join=round,line cap=round,fill=fillColor,fill opacity=0.80] ( 89.84,208.54) circle (  0.89);
\definecolor{drawColor}{gray}{0.20}

\path[draw=drawColor,line width= 0.6pt,line join=round] ( 89.84,230.98) -- ( 89.84,241.29);

\path[draw=drawColor,line width= 0.6pt,line join=round] ( 89.84,222.77) -- ( 89.84,210.83);
\definecolor{fillColor}{RGB}{68,119,170}

\path[draw=drawColor,line width= 0.6pt,line join=round,line cap=round,fill=fillColor,fill opacity=0.80] ( 86.06,230.98) --
	( 86.06,222.77) --
	( 93.62,222.77) --
	( 93.62,230.98) --
	( 86.06,230.98) --
	cycle;

\path[draw=drawColor,line width= 1.1pt,line join=round] ( 86.06,229.48) -- ( 93.62,229.48);

\path[draw=drawColor,line width= 0.6pt,line join=round] ( 98.24,230.37) -- ( 98.24,235.97);

\path[draw=drawColor,line width= 0.6pt,line join=round] ( 98.24,217.87) -- ( 98.24,208.15);
\definecolor{fillColor}{RGB}{204,102,119}

\path[draw=drawColor,line width= 0.6pt,line join=round,line cap=round,fill=fillColor,fill opacity=0.80] ( 94.46,230.37) --
	( 94.46,217.87) --
	(102.02,217.87) --
	(102.02,230.37) --
	( 94.46,230.37) --
	cycle;

\path[draw=drawColor,line width= 1.1pt,line join=round] ( 94.46,224.06) -- (102.02,224.06);

\path[draw=drawColor,line width= 0.6pt,line join=round] (112.24,222.84) -- (112.24,222.84);

\path[draw=drawColor,line width= 0.6pt,line join=round] (112.24,222.84) -- (112.24,222.84);
\definecolor{fillColor}{RGB}{68,119,170}

\path[draw=drawColor,line width= 0.6pt,line join=round,line cap=round,fill=fillColor,fill opacity=0.80] (108.46,222.84) --
	(108.46,222.84) --
	(116.01,222.84) --
	(116.01,222.84) --
	(108.46,222.84) --
	cycle;

\path[draw=drawColor,line width= 1.1pt,line join=round] (108.46,222.84) -- (116.01,222.84);

\path[draw=drawColor,line width= 0.6pt,line join=round] (120.63,220.86) -- (120.63,220.86);

\path[draw=drawColor,line width= 0.6pt,line join=round] (120.63,220.86) -- (120.63,220.86);
\definecolor{fillColor}{RGB}{204,102,119}

\path[draw=drawColor,line width= 0.6pt,line join=round,line cap=round,fill=fillColor,fill opacity=0.80] (116.85,220.86) --
	(116.85,220.86) --
	(124.41,220.86) --
	(124.41,220.86) --
	(116.85,220.86) --
	cycle;

\path[draw=drawColor,line width= 1.1pt,line join=round] (116.85,220.86) -- (124.41,220.86);

\path[draw=drawColor,line width= 0.6pt,line join=round] (134.63,229.30) -- (134.63,230.39);

\path[draw=drawColor,line width= 0.6pt,line join=round] (134.63,220.72) -- (134.63,219.10);
\definecolor{fillColor}{RGB}{68,119,170}

\path[draw=drawColor,line width= 0.6pt,line join=round,line cap=round,fill=fillColor,fill opacity=0.80] (130.85,229.30) --
	(130.85,220.72) --
	(138.41,220.72) --
	(138.41,229.30) --
	(130.85,229.30) --
	cycle;

\path[draw=drawColor,line width= 1.1pt,line join=round] (130.85,225.50) -- (138.41,225.50);

\path[draw=drawColor,line width= 0.6pt,line join=round] (143.03,227.11) -- (143.03,228.42);

\path[draw=drawColor,line width= 0.6pt,line join=round] (143.03,223.83) -- (143.03,223.40);
\definecolor{fillColor}{RGB}{204,102,119}

\path[draw=drawColor,line width= 0.6pt,line join=round,line cap=round,fill=fillColor,fill opacity=0.80] (139.25,227.11) --
	(139.25,223.83) --
	(146.81,223.83) --
	(146.81,227.11) --
	(139.25,227.11) --
	cycle;

\path[draw=drawColor,line width= 1.1pt,line join=round] (139.25,224.71) -- (146.81,224.71);

\path[draw=drawColor,line width= 0.6pt,line join=round] (157.02,234.47) -- (157.02,239.62);

\path[draw=drawColor,line width= 0.6pt,line join=round] (157.02,225.39) -- (157.02,213.69);
\definecolor{fillColor}{RGB}{68,119,170}

\path[draw=drawColor,line width= 0.6pt,line join=round,line cap=round,fill=fillColor,fill opacity=0.80] (153.24,234.47) --
	(153.24,225.39) --
	(160.80,225.39) --
	(160.80,234.47) --
	(153.24,234.47) --
	cycle;

\path[draw=drawColor,line width= 1.1pt,line join=round] (153.24,232.37) -- (160.80,232.37);

\path[draw=drawColor,line width= 0.6pt,line join=round] (165.42,230.00) -- (165.42,231.67);

\path[draw=drawColor,line width= 0.6pt,line join=round] (165.42,219.03) -- (165.42,207.66);
\definecolor{fillColor}{RGB}{204,102,119}

\path[draw=drawColor,line width= 0.6pt,line join=round,line cap=round,fill=fillColor,fill opacity=0.80] (161.64,230.00) --
	(161.64,219.03) --
	(169.20,219.03) --
	(169.20,230.00) --
	(161.64,230.00) --
	cycle;

\path[draw=drawColor,line width= 1.1pt,line join=round] (161.64,226.58) -- (169.20,226.58);

\path[draw=drawColor,line width= 0.6pt,line join=round] (179.42,229.26) -- (179.42,229.48);

\path[draw=drawColor,line width= 0.6pt,line join=round] (179.42,222.79) -- (179.42,214.06);
\definecolor{fillColor}{RGB}{68,119,170}

\path[draw=drawColor,line width= 0.6pt,line join=round,line cap=round,fill=fillColor,fill opacity=0.80] (175.64,229.26) --
	(175.64,222.79) --
	(183.20,222.79) --
	(183.20,229.26) --
	(175.64,229.26) --
	cycle;

\path[draw=drawColor,line width= 1.1pt,line join=round] (175.64,228.92) -- (183.20,228.92);

\path[draw=drawColor,line width= 0.6pt,line join=round] (187.82,230.22) -- (187.82,230.80);

\path[draw=drawColor,line width= 0.6pt,line join=round] (187.82,226.03) -- (187.82,223.49);
\definecolor{fillColor}{RGB}{204,102,119}

\path[draw=drawColor,line width= 0.6pt,line join=round,line cap=round,fill=fillColor,fill opacity=0.80] (184.04,230.22) --
	(184.04,226.03) --
	(191.59,226.03) --
	(191.59,230.22) --
	(184.04,230.22) --
	cycle;

\path[draw=drawColor,line width= 1.1pt,line join=round] (184.04,229.21) -- (191.59,229.21);
\end{scope}
\begin{scope}
\path[clip] ( 58.21,104.05) rectangle (197.05,174.26);
\definecolor{fillColor}{RGB}{255,255,255}

\path[fill=fillColor] ( 58.21,104.05) rectangle (197.05,174.26);
\definecolor{drawColor}{RGB}{51,51,51}
\definecolor{fillColor}{RGB}{51,51,51}

\path[draw=drawColor,draw opacity=0.80,line width= 0.4pt,line join=round,line cap=round,fill=fillColor,fill opacity=0.80] ( 67.45,165.69) circle (  0.89);

\path[draw=drawColor,draw opacity=0.80,line width= 0.4pt,line join=round,line cap=round,fill=fillColor,fill opacity=0.80] ( 67.45,171.07) circle (  0.89);

\path[draw=drawColor,draw opacity=0.80,line width= 0.4pt,line join=round,line cap=round,fill=fillColor,fill opacity=0.80] ( 67.45,151.98) circle (  0.89);
\definecolor{drawColor}{gray}{0.20}

\path[draw=drawColor,line width= 0.6pt,line join=round] ( 67.45,133.61) -- ( 67.45,150.26);

\path[draw=drawColor,line width= 0.6pt,line join=round] ( 67.45,121.86) -- ( 67.45,111.06);
\definecolor{fillColor}{RGB}{68,119,170}

\path[draw=drawColor,line width= 0.6pt,line join=round,line cap=round,fill=fillColor,fill opacity=0.80] ( 63.67,133.61) --
	( 63.67,121.86) --
	( 71.23,121.86) --
	( 71.23,133.61) --
	( 63.67,133.61) --
	cycle;

\path[draw=drawColor,line width= 1.1pt,line join=round] ( 63.67,125.89) -- ( 71.23,125.89);

\path[draw=drawColor,line width= 0.6pt,line join=round] ( 75.85,129.25) -- ( 75.85,141.22);

\path[draw=drawColor,line width= 0.6pt,line join=round] ( 75.85,118.08) -- ( 75.85,107.24);
\definecolor{fillColor}{RGB}{204,102,119}

\path[draw=drawColor,line width= 0.6pt,line join=round,line cap=round,fill=fillColor,fill opacity=0.80] ( 72.07,129.25) --
	( 72.07,118.08) --
	( 79.62,118.08) --
	( 79.62,129.25) --
	( 72.07,129.25) --
	cycle;

\path[draw=drawColor,line width= 1.1pt,line join=round] ( 72.07,124.24) -- ( 79.62,124.24);

\path[draw=drawColor,line width= 0.6pt,line join=round] ( 89.84,123.47) -- ( 89.84,125.91);

\path[draw=drawColor,line width= 0.6pt,line join=round] ( 89.84,118.57) -- ( 89.84,116.13);
\definecolor{fillColor}{RGB}{68,119,170}

\path[draw=drawColor,line width= 0.6pt,line join=round,line cap=round,fill=fillColor,fill opacity=0.80] ( 86.06,123.47) --
	( 86.06,118.57) --
	( 93.62,118.57) --
	( 93.62,123.47) --
	( 86.06,123.47) --
	cycle;

\path[draw=drawColor,line width= 1.1pt,line join=round] ( 86.06,121.02) -- ( 93.62,121.02);

\path[draw=drawColor,line width= 0.6pt,line join=round] ( 98.24,123.75) -- ( 98.24,124.22);

\path[draw=drawColor,line width= 0.6pt,line join=round] ( 98.24,122.80) -- ( 98.24,122.33);
\definecolor{fillColor}{RGB}{204,102,119}

\path[draw=drawColor,line width= 0.6pt,line join=round,line cap=round,fill=fillColor,fill opacity=0.80] ( 94.46,123.75) --
	( 94.46,122.80) --
	(102.02,122.80) --
	(102.02,123.75) --
	( 94.46,123.75) --
	cycle;

\path[draw=drawColor,line width= 1.1pt,line join=round] ( 94.46,123.28) -- (102.02,123.28);

\path[draw=drawColor,line width= 0.6pt,line join=round] (112.24,125.69) -- (112.24,125.69);

\path[draw=drawColor,line width= 0.6pt,line join=round] (112.24,125.69) -- (112.24,125.69);
\definecolor{fillColor}{RGB}{68,119,170}

\path[draw=drawColor,line width= 0.6pt,line join=round,line cap=round,fill=fillColor,fill opacity=0.80] (108.46,125.69) --
	(108.46,125.69) --
	(116.01,125.69) --
	(116.01,125.69) --
	(108.46,125.69) --
	cycle;

\path[draw=drawColor,line width= 1.1pt,line join=round] (108.46,125.69) -- (116.01,125.69);

\path[draw=drawColor,line width= 0.6pt,line join=round] (120.63,126.85) -- (120.63,126.85);

\path[draw=drawColor,line width= 0.6pt,line join=round] (120.63,126.85) -- (120.63,126.85);
\definecolor{fillColor}{RGB}{204,102,119}

\path[draw=drawColor,line width= 0.6pt,line join=round,line cap=round,fill=fillColor,fill opacity=0.80] (116.85,126.85) --
	(116.85,126.85) --
	(124.41,126.85) --
	(124.41,126.85) --
	(116.85,126.85) --
	cycle;

\path[draw=drawColor,line width= 1.1pt,line join=round] (116.85,126.85) -- (124.41,126.85);
\definecolor{drawColor}{RGB}{51,51,51}
\definecolor{fillColor}{RGB}{51,51,51}

\path[draw=drawColor,draw opacity=0.80,line width= 0.4pt,line join=round,line cap=round,fill=fillColor,fill opacity=0.80] (134.63,136.50) circle (  0.89);
\definecolor{drawColor}{gray}{0.20}

\path[draw=drawColor,line width= 0.6pt,line join=round] (134.63,131.88) -- (134.63,132.24);

\path[draw=drawColor,line width= 0.6pt,line join=round] (134.63,130.23) -- (134.63,129.64);
\definecolor{fillColor}{RGB}{68,119,170}

\path[draw=drawColor,line width= 0.6pt,line join=round,line cap=round,fill=fillColor,fill opacity=0.80] (130.85,131.88) --
	(130.85,130.23) --
	(138.41,130.23) --
	(138.41,131.88) --
	(130.85,131.88) --
	cycle;

\path[draw=drawColor,line width= 1.1pt,line join=round] (130.85,131.01) -- (138.41,131.01);

\path[draw=drawColor,line width= 0.6pt,line join=round] (143.03,137.42) -- (143.03,138.52);

\path[draw=drawColor,line width= 0.6pt,line join=round] (143.03,134.24) -- (143.03,133.59);
\definecolor{fillColor}{RGB}{204,102,119}

\path[draw=drawColor,line width= 0.6pt,line join=round,line cap=round,fill=fillColor,fill opacity=0.80] (139.25,137.42) --
	(139.25,134.24) --
	(146.81,134.24) --
	(146.81,137.42) --
	(139.25,137.42) --
	cycle;

\path[draw=drawColor,line width= 1.1pt,line join=round] (139.25,135.79) -- (146.81,135.79);

\path[draw=drawColor,line width= 0.6pt,line join=round] (179.42,124.07) -- (179.42,129.87);

\path[draw=drawColor,line width= 0.6pt,line join=round] (179.42,112.79) -- (179.42,108.65);
\definecolor{fillColor}{RGB}{68,119,170}

\path[draw=drawColor,line width= 0.6pt,line join=round,line cap=round,fill=fillColor,fill opacity=0.80] (175.64,124.07) --
	(175.64,112.79) --
	(183.20,112.79) --
	(183.20,124.07) --
	(175.64,124.07) --
	cycle;

\path[draw=drawColor,line width= 1.1pt,line join=round] (175.64,116.02) -- (183.20,116.02);
\definecolor{drawColor}{RGB}{51,51,51}
\definecolor{fillColor}{RGB}{51,51,51}

\path[draw=drawColor,draw opacity=0.80,line width= 0.4pt,line join=round,line cap=round,fill=fillColor,fill opacity=0.80] (187.82,123.01) circle (  0.89);
\definecolor{drawColor}{gray}{0.20}

\path[draw=drawColor,line width= 0.6pt,line join=round] (187.82,137.44) -- (187.82,142.73);

\path[draw=drawColor,line width= 0.6pt,line join=round] (187.82,132.01) -- (187.82,125.89);
\definecolor{fillColor}{RGB}{204,102,119}

\path[draw=drawColor,line width= 0.6pt,line join=round,line cap=round,fill=fillColor,fill opacity=0.80] (184.04,137.44) --
	(184.04,132.01) --
	(191.59,132.01) --
	(191.59,137.44) --
	(184.04,137.44) --
	cycle;

\path[draw=drawColor,line width= 1.1pt,line join=round] (184.04,136.01) -- (191.59,136.01);
\end{scope}
\begin{scope}
\path[clip] (202.55,196.56) rectangle (341.40,266.78);
\definecolor{fillColor}{RGB}{255,255,255}

\path[fill=fillColor] (202.55,196.56) rectangle (341.40,266.78);
\definecolor{drawColor}{gray}{0.20}

\path[draw=drawColor,line width= 0.6pt,line join=round] (211.79,238.16) -- (211.79,243.81);

\path[draw=drawColor,line width= 0.6pt,line join=round] (211.79,228.78) -- (211.79,226.34);
\definecolor{fillColor}{RGB}{68,119,170}

\path[draw=drawColor,line width= 0.6pt,line join=round,line cap=round,fill=fillColor,fill opacity=0.80] (208.01,238.16) --
	(208.01,228.78) --
	(215.57,228.78) --
	(215.57,238.16) --
	(208.01,238.16) --
	cycle;

\path[draw=drawColor,line width= 1.1pt,line join=round] (208.01,236.26) -- (215.57,236.26);

\path[draw=drawColor,line width= 0.6pt,line join=round] (220.19,231.25) -- (220.19,237.15);

\path[draw=drawColor,line width= 0.6pt,line join=round] (220.19,215.84) -- (220.19,208.51);
\definecolor{fillColor}{RGB}{204,102,119}

\path[draw=drawColor,line width= 0.6pt,line join=round,line cap=round,fill=fillColor,fill opacity=0.80] (216.41,231.25) --
	(216.41,215.84) --
	(223.97,215.84) --
	(223.97,231.25) --
	(216.41,231.25) --
	cycle;

\path[draw=drawColor,line width= 1.1pt,line join=round] (216.41,226.19) -- (223.97,226.19);
\definecolor{drawColor}{RGB}{51,51,51}
\definecolor{fillColor}{RGB}{51,51,51}

\path[draw=drawColor,draw opacity=0.80,line width= 0.4pt,line join=round,line cap=round,fill=fillColor,fill opacity=0.80] (234.18,239.11) circle (  0.89);

\path[draw=drawColor,draw opacity=0.80,line width= 0.4pt,line join=round,line cap=round,fill=fillColor,fill opacity=0.80] (234.18,217.95) circle (  0.89);

\path[draw=drawColor,draw opacity=0.80,line width= 0.4pt,line join=round,line cap=round,fill=fillColor,fill opacity=0.80] (234.18,219.37) circle (  0.89);

\path[draw=drawColor,draw opacity=0.80,line width= 0.4pt,line join=round,line cap=round,fill=fillColor,fill opacity=0.80] (234.18,218.03) circle (  0.89);

\path[draw=drawColor,draw opacity=0.80,line width= 0.4pt,line join=round,line cap=round,fill=fillColor,fill opacity=0.80] (234.18,221.05) circle (  0.89);

\path[draw=drawColor,draw opacity=0.80,line width= 0.4pt,line join=round,line cap=round,fill=fillColor,fill opacity=0.80] (234.18,218.79) circle (  0.89);

\path[draw=drawColor,draw opacity=0.80,line width= 0.4pt,line join=round,line cap=round,fill=fillColor,fill opacity=0.80] (234.18,239.24) circle (  0.89);

\path[draw=drawColor,draw opacity=0.80,line width= 0.4pt,line join=round,line cap=round,fill=fillColor,fill opacity=0.80] (234.18,240.89) circle (  0.89);

\path[draw=drawColor,draw opacity=0.80,line width= 0.4pt,line join=round,line cap=round,fill=fillColor,fill opacity=0.80] (234.18,219.02) circle (  0.89);

\path[draw=drawColor,draw opacity=0.80,line width= 0.4pt,line join=round,line cap=round,fill=fillColor,fill opacity=0.80] (234.18,240.61) circle (  0.89);

\path[draw=drawColor,draw opacity=0.80,line width= 0.4pt,line join=round,line cap=round,fill=fillColor,fill opacity=0.80] (234.18,236.91) circle (  0.89);

\path[draw=drawColor,draw opacity=0.80,line width= 0.4pt,line join=round,line cap=round,fill=fillColor,fill opacity=0.80] (234.18,245.00) circle (  0.89);

\path[draw=drawColor,draw opacity=0.80,line width= 0.4pt,line join=round,line cap=round,fill=fillColor,fill opacity=0.80] (234.18,217.67) circle (  0.89);

\path[draw=drawColor,draw opacity=0.80,line width= 0.4pt,line join=round,line cap=round,fill=fillColor,fill opacity=0.80] (234.18,221.01) circle (  0.89);

\path[draw=drawColor,draw opacity=0.80,line width= 0.4pt,line join=round,line cap=round,fill=fillColor,fill opacity=0.80] (234.18,214.48) circle (  0.89);

\path[draw=drawColor,draw opacity=0.80,line width= 0.4pt,line join=round,line cap=round,fill=fillColor,fill opacity=0.80] (234.18,239.72) circle (  0.89);

\path[draw=drawColor,draw opacity=0.80,line width= 0.4pt,line join=round,line cap=round,fill=fillColor,fill opacity=0.80] (234.18,220.54) circle (  0.89);

\path[draw=drawColor,draw opacity=0.80,line width= 0.4pt,line join=round,line cap=round,fill=fillColor,fill opacity=0.80] (234.18,247.86) circle (  0.89);

\path[draw=drawColor,draw opacity=0.80,line width= 0.4pt,line join=round,line cap=round,fill=fillColor,fill opacity=0.80] (234.18,212.68) circle (  0.89);

\path[draw=drawColor,draw opacity=0.80,line width= 0.4pt,line join=round,line cap=round,fill=fillColor,fill opacity=0.80] (234.18,242.81) circle (  0.89);

\path[draw=drawColor,draw opacity=0.80,line width= 0.4pt,line join=round,line cap=round,fill=fillColor,fill opacity=0.80] (234.18,243.80) circle (  0.89);
\definecolor{drawColor}{gray}{0.20}

\path[draw=drawColor,line width= 0.6pt,line join=round] (234.18,230.98) -- (234.18,236.66);

\path[draw=drawColor,line width= 0.6pt,line join=round] (234.18,227.08) -- (234.18,221.51);
\definecolor{fillColor}{RGB}{68,119,170}

\path[draw=drawColor,line width= 0.6pt,line join=round,line cap=round,fill=fillColor,fill opacity=0.80] (230.41,230.98) --
	(230.41,227.08) --
	(237.96,227.08) --
	(237.96,230.98) --
	(230.41,230.98) --
	cycle;

\path[draw=drawColor,line width= 1.1pt,line join=round] (230.41,230.98) -- (237.96,230.98);

\path[draw=drawColor,line width= 0.6pt,line join=round] (242.58,234.08) -- (242.58,246.24);

\path[draw=drawColor,line width= 0.6pt,line join=round] (242.58,225.71) -- (242.58,214.92);
\definecolor{fillColor}{RGB}{204,102,119}

\path[draw=drawColor,line width= 0.6pt,line join=round,line cap=round,fill=fillColor,fill opacity=0.80] (238.80,234.08) --
	(238.80,225.71) --
	(246.36,225.71) --
	(246.36,234.08) --
	(238.80,234.08) --
	cycle;

\path[draw=drawColor,line width= 1.1pt,line join=round] (238.80,229.17) -- (246.36,229.17);

\path[draw=drawColor,line width= 0.6pt,line join=round] (256.58,228.00) -- (256.58,228.00);

\path[draw=drawColor,line width= 0.6pt,line join=round] (256.58,228.00) -- (256.58,228.00);
\definecolor{fillColor}{RGB}{68,119,170}

\path[draw=drawColor,line width= 0.6pt,line join=round,line cap=round,fill=fillColor,fill opacity=0.80] (252.80,228.00) --
	(252.80,228.00) --
	(260.36,228.00) --
	(260.36,228.00) --
	(252.80,228.00) --
	cycle;

\path[draw=drawColor,line width= 1.1pt,line join=round] (252.80,228.00) -- (260.36,228.00);

\path[draw=drawColor,line width= 0.6pt,line join=round] (264.98,226.99) -- (264.98,226.99);

\path[draw=drawColor,line width= 0.6pt,line join=round] (264.98,226.99) -- (264.98,226.99);
\definecolor{fillColor}{RGB}{204,102,119}

\path[draw=drawColor,line width= 0.6pt,line join=round,line cap=round,fill=fillColor,fill opacity=0.80] (261.20,226.99) --
	(261.20,226.99) --
	(268.76,226.99) --
	(268.76,226.99) --
	(261.20,226.99) --
	cycle;

\path[draw=drawColor,line width= 1.1pt,line join=round] (261.20,226.99) -- (268.76,226.99);

\path[draw=drawColor,line width= 0.6pt,line join=round] (278.97,239.17) -- (278.97,242.14);

\path[draw=drawColor,line width= 0.6pt,line join=round] (278.97,230.78) -- (278.97,229.24);
\definecolor{fillColor}{RGB}{68,119,170}

\path[draw=drawColor,line width= 0.6pt,line join=round,line cap=round,fill=fillColor,fill opacity=0.80] (275.19,239.17) --
	(275.19,230.78) --
	(282.75,230.78) --
	(282.75,239.17) --
	(275.19,239.17) --
	cycle;

\path[draw=drawColor,line width= 1.1pt,line join=round] (275.19,232.89) -- (282.75,232.89);
\definecolor{drawColor}{RGB}{51,51,51}
\definecolor{fillColor}{RGB}{51,51,51}

\path[draw=drawColor,draw opacity=0.80,line width= 0.4pt,line join=round,line cap=round,fill=fillColor,fill opacity=0.80] (287.37,232.90) circle (  0.89);

\path[draw=drawColor,draw opacity=0.80,line width= 0.4pt,line join=round,line cap=round,fill=fillColor,fill opacity=0.80] (287.37,227.57) circle (  0.89);
\definecolor{drawColor}{gray}{0.20}

\path[draw=drawColor,line width= 0.6pt,line join=round] (287.37,230.98) -- (287.37,230.98);

\path[draw=drawColor,line width= 0.6pt,line join=round] (287.37,230.98) -- (287.37,230.98);
\definecolor{fillColor}{RGB}{204,102,119}

\path[draw=drawColor,line width= 0.6pt,line join=round,line cap=round,fill=fillColor,fill opacity=0.80] (283.59,230.98) --
	(283.59,230.98) --
	(291.15,230.98) --
	(291.15,230.98) --
	(283.59,230.98) --
	cycle;

\path[draw=drawColor,line width= 1.1pt,line join=round] (283.59,230.98) -- (291.15,230.98);
\definecolor{drawColor}{RGB}{51,51,51}
\definecolor{fillColor}{RGB}{51,51,51}

\path[draw=drawColor,draw opacity=0.80,line width= 0.4pt,line join=round,line cap=round,fill=fillColor,fill opacity=0.80] (301.37,224.80) circle (  0.89);
\definecolor{drawColor}{gray}{0.20}

\path[draw=drawColor,line width= 0.6pt,line join=round] (301.37,236.58) -- (301.37,241.68);

\path[draw=drawColor,line width= 0.6pt,line join=round] (301.37,231.91) -- (301.37,225.39);
\definecolor{fillColor}{RGB}{68,119,170}

\path[draw=drawColor,line width= 0.6pt,line join=round,line cap=round,fill=fillColor,fill opacity=0.80] (297.59,236.58) --
	(297.59,231.91) --
	(305.15,231.91) --
	(305.15,236.58) --
	(297.59,236.58) --
	cycle;

\path[draw=drawColor,line width= 1.1pt,line join=round] (297.59,233.84) -- (305.15,233.84);
\definecolor{drawColor}{RGB}{51,51,51}
\definecolor{fillColor}{RGB}{51,51,51}

\path[draw=drawColor,draw opacity=0.80,line width= 0.4pt,line join=round,line cap=round,fill=fillColor,fill opacity=0.80] (309.76,216.27) circle (  0.89);

\path[draw=drawColor,draw opacity=0.80,line width= 0.4pt,line join=round,line cap=round,fill=fillColor,fill opacity=0.80] (309.76,221.13) circle (  0.89);

\path[draw=drawColor,draw opacity=0.80,line width= 0.4pt,line join=round,line cap=round,fill=fillColor,fill opacity=0.80] (309.76,218.23) circle (  0.89);

\path[draw=drawColor,draw opacity=0.80,line width= 0.4pt,line join=round,line cap=round,fill=fillColor,fill opacity=0.80] (309.76,217.97) circle (  0.89);

\path[draw=drawColor,draw opacity=0.80,line width= 0.4pt,line join=round,line cap=round,fill=fillColor,fill opacity=0.80] (309.76,212.67) circle (  0.89);

\path[draw=drawColor,draw opacity=0.80,line width= 0.4pt,line join=round,line cap=round,fill=fillColor,fill opacity=0.80] (309.76,221.84) circle (  0.89);

\path[draw=drawColor,draw opacity=0.80,line width= 0.4pt,line join=round,line cap=round,fill=fillColor,fill opacity=0.80] (309.76,235.57) circle (  0.89);

\path[draw=drawColor,draw opacity=0.80,line width= 0.4pt,line join=round,line cap=round,fill=fillColor,fill opacity=0.80] (309.76,217.94) circle (  0.89);

\path[draw=drawColor,draw opacity=0.80,line width= 0.4pt,line join=round,line cap=round,fill=fillColor,fill opacity=0.80] (309.76,235.60) circle (  0.89);

\path[draw=drawColor,draw opacity=0.80,line width= 0.4pt,line join=round,line cap=round,fill=fillColor,fill opacity=0.80] (309.76,219.99) circle (  0.89);

\path[draw=drawColor,draw opacity=0.80,line width= 0.4pt,line join=round,line cap=round,fill=fillColor,fill opacity=0.80] (309.76,218.22) circle (  0.89);

\path[draw=drawColor,draw opacity=0.80,line width= 0.4pt,line join=round,line cap=round,fill=fillColor,fill opacity=0.80] (309.76,218.16) circle (  0.89);

\path[draw=drawColor,draw opacity=0.80,line width= 0.4pt,line join=round,line cap=round,fill=fillColor,fill opacity=0.80] (309.76,220.50) circle (  0.89);

\path[draw=drawColor,draw opacity=0.80,line width= 0.4pt,line join=round,line cap=round,fill=fillColor,fill opacity=0.80] (309.76,219.92) circle (  0.89);

\path[draw=drawColor,draw opacity=0.80,line width= 0.4pt,line join=round,line cap=round,fill=fillColor,fill opacity=0.80] (309.76,219.45) circle (  0.89);

\path[draw=drawColor,draw opacity=0.80,line width= 0.4pt,line join=round,line cap=round,fill=fillColor,fill opacity=0.80] (309.76,221.06) circle (  0.89);

\path[draw=drawColor,draw opacity=0.80,line width= 0.4pt,line join=round,line cap=round,fill=fillColor,fill opacity=0.80] (309.76,221.33) circle (  0.89);
\definecolor{drawColor}{gray}{0.20}

\path[draw=drawColor,line width= 0.6pt,line join=round] (309.76,230.49) -- (309.76,235.21);

\path[draw=drawColor,line width= 0.6pt,line join=round] (309.76,227.15) -- (309.76,222.30);
\definecolor{fillColor}{RGB}{204,102,119}

\path[draw=drawColor,line width= 0.6pt,line join=round,line cap=round,fill=fillColor,fill opacity=0.80] (305.99,230.49) --
	(305.99,227.15) --
	(313.54,227.15) --
	(313.54,230.49) --
	(305.99,230.49) --
	cycle;

\path[draw=drawColor,line width= 1.1pt,line join=round] (305.99,229.31) -- (313.54,229.31);

\path[draw=drawColor,line width= 0.6pt,line join=round] (323.76,224.15) -- (323.76,227.80);

\path[draw=drawColor,line width= 0.6pt,line join=round] (323.76,221.51) -- (323.76,219.55);
\definecolor{fillColor}{RGB}{68,119,170}

\path[draw=drawColor,line width= 0.6pt,line join=round,line cap=round,fill=fillColor,fill opacity=0.80] (319.98,224.15) --
	(319.98,221.51) --
	(327.54,221.51) --
	(327.54,224.15) --
	(319.98,224.15) --
	cycle;

\path[draw=drawColor,line width= 1.1pt,line join=round] (319.98,222.65) -- (327.54,222.65);
\definecolor{drawColor}{RGB}{51,51,51}
\definecolor{fillColor}{RGB}{51,51,51}

\path[draw=drawColor,draw opacity=0.80,line width= 0.4pt,line join=round,line cap=round,fill=fillColor,fill opacity=0.80] (332.16,233.13) circle (  0.89);

\path[draw=drawColor,draw opacity=0.80,line width= 0.4pt,line join=round,line cap=round,fill=fillColor,fill opacity=0.80] (332.16,232.21) circle (  0.89);

\path[draw=drawColor,draw opacity=0.80,line width= 0.4pt,line join=round,line cap=round,fill=fillColor,fill opacity=0.80] (332.16,226.27) circle (  0.89);
\definecolor{drawColor}{gray}{0.20}

\path[draw=drawColor,line width= 0.6pt,line join=round] (332.16,230.78) -- (332.16,230.91);

\path[draw=drawColor,line width= 0.6pt,line join=round] (332.16,229.91) -- (332.16,229.17);
\definecolor{fillColor}{RGB}{204,102,119}

\path[draw=drawColor,line width= 0.6pt,line join=round,line cap=round,fill=fillColor,fill opacity=0.80] (328.38,230.78) --
	(328.38,229.91) --
	(335.94,229.91) --
	(335.94,230.78) --
	(328.38,230.78) --
	cycle;

\path[draw=drawColor,line width= 1.1pt,line join=round] (328.38,230.55) -- (335.94,230.55);
\end{scope}
\begin{scope}
\path[clip] (202.55,104.05) rectangle (341.40,174.26);
\definecolor{fillColor}{RGB}{255,255,255}

\path[fill=fillColor] (202.55,104.05) rectangle (341.40,174.26);
\definecolor{drawColor}{gray}{0.20}

\path[draw=drawColor,line width= 0.6pt,line join=round] (234.18,137.99) -- (234.18,152.36);

\path[draw=drawColor,line width= 0.6pt,line join=round] (234.18,128.08) -- (234.18,121.49);
\definecolor{fillColor}{RGB}{68,119,170}

\path[draw=drawColor,line width= 0.6pt,line join=round,line cap=round,fill=fillColor,fill opacity=0.80] (230.41,137.99) --
	(230.41,128.08) --
	(237.96,128.08) --
	(237.96,137.99) --
	(230.41,137.99) --
	cycle;

\path[draw=drawColor,line width= 1.1pt,line join=round] (230.41,132.42) -- (237.96,132.42);
\definecolor{drawColor}{RGB}{51,51,51}
\definecolor{fillColor}{RGB}{51,51,51}

\path[draw=drawColor,draw opacity=0.80,line width= 0.4pt,line join=round,line cap=round,fill=fillColor,fill opacity=0.80] (242.58,144.19) circle (  0.89);

\path[draw=drawColor,draw opacity=0.80,line width= 0.4pt,line join=round,line cap=round,fill=fillColor,fill opacity=0.80] (242.58,150.91) circle (  0.89);

\path[draw=drawColor,draw opacity=0.80,line width= 0.4pt,line join=round,line cap=round,fill=fillColor,fill opacity=0.80] (242.58,152.42) circle (  0.89);

\path[draw=drawColor,draw opacity=0.80,line width= 0.4pt,line join=round,line cap=round,fill=fillColor,fill opacity=0.80] (242.58,146.87) circle (  0.89);

\path[draw=drawColor,draw opacity=0.80,line width= 0.4pt,line join=round,line cap=round,fill=fillColor,fill opacity=0.80] (242.58,145.36) circle (  0.89);

\path[draw=drawColor,draw opacity=0.80,line width= 0.4pt,line join=round,line cap=round,fill=fillColor,fill opacity=0.80] (242.58,148.93) circle (  0.89);

\path[draw=drawColor,draw opacity=0.80,line width= 0.4pt,line join=round,line cap=round,fill=fillColor,fill opacity=0.80] (242.58,115.36) circle (  0.89);
\definecolor{drawColor}{gray}{0.20}

\path[draw=drawColor,line width= 0.6pt,line join=round] (242.58,134.02) -- (242.58,143.72);

\path[draw=drawColor,line width= 0.6pt,line join=round] (242.58,127.34) -- (242.58,118.37);
\definecolor{fillColor}{RGB}{204,102,119}

\path[draw=drawColor,line width= 0.6pt,line join=round,line cap=round,fill=fillColor,fill opacity=0.80] (238.80,134.02) --
	(238.80,127.34) --
	(246.36,127.34) --
	(246.36,134.02) --
	(238.80,134.02) --
	cycle;

\path[draw=drawColor,line width= 1.1pt,line join=round] (238.80,130.34) -- (246.36,130.34);
\definecolor{drawColor}{RGB}{51,51,51}
\definecolor{fillColor}{RGB}{51,51,51}

\path[draw=drawColor,draw opacity=0.80,line width= 0.4pt,line join=round,line cap=round,fill=fillColor,fill opacity=0.80] (301.37,122.69) circle (  0.89);

\path[draw=drawColor,draw opacity=0.80,line width= 0.4pt,line join=round,line cap=round,fill=fillColor,fill opacity=0.80] (301.37,122.16) circle (  0.89);

\path[draw=drawColor,draw opacity=0.80,line width= 0.4pt,line join=round,line cap=round,fill=fillColor,fill opacity=0.80] (301.37,119.60) circle (  0.89);

\path[draw=drawColor,draw opacity=0.80,line width= 0.4pt,line join=round,line cap=round,fill=fillColor,fill opacity=0.80] (301.37,121.31) circle (  0.89);

\path[draw=drawColor,draw opacity=0.80,line width= 0.4pt,line join=round,line cap=round,fill=fillColor,fill opacity=0.80] (301.37,113.13) circle (  0.89);

\path[draw=drawColor,draw opacity=0.80,line width= 0.4pt,line join=round,line cap=round,fill=fillColor,fill opacity=0.80] (301.37,147.00) circle (  0.89);
\definecolor{drawColor}{gray}{0.20}

\path[draw=drawColor,line width= 0.6pt,line join=round] (301.37,136.71) -- (301.37,143.53);

\path[draw=drawColor,line width= 0.6pt,line join=round] (301.37,131.18) -- (301.37,123.15);
\definecolor{fillColor}{RGB}{68,119,170}

\path[draw=drawColor,line width= 0.6pt,line join=round,line cap=round,fill=fillColor,fill opacity=0.80] (297.59,136.71) --
	(297.59,131.18) --
	(305.15,131.18) --
	(305.15,136.71) --
	(297.59,136.71) --
	cycle;

\path[draw=drawColor,line width= 1.1pt,line join=round] (297.59,133.96) -- (305.15,133.96);
\definecolor{drawColor}{RGB}{51,51,51}
\definecolor{fillColor}{RGB}{51,51,51}

\path[draw=drawColor,draw opacity=0.80,line width= 0.4pt,line join=round,line cap=round,fill=fillColor,fill opacity=0.80] (309.76,119.41) circle (  0.89);

\path[draw=drawColor,draw opacity=0.80,line width= 0.4pt,line join=round,line cap=round,fill=fillColor,fill opacity=0.80] (309.76,122.44) circle (  0.89);

\path[draw=drawColor,draw opacity=0.80,line width= 0.4pt,line join=round,line cap=round,fill=fillColor,fill opacity=0.80] (309.76,118.18) circle (  0.89);

\path[draw=drawColor,draw opacity=0.80,line width= 0.4pt,line join=round,line cap=round,fill=fillColor,fill opacity=0.80] (309.76,122.26) circle (  0.89);

\path[draw=drawColor,draw opacity=0.80,line width= 0.4pt,line join=round,line cap=round,fill=fillColor,fill opacity=0.80] (309.76,116.23) circle (  0.89);

\path[draw=drawColor,draw opacity=0.80,line width= 0.4pt,line join=round,line cap=round,fill=fillColor,fill opacity=0.80] (309.76,122.05) circle (  0.89);

\path[draw=drawColor,draw opacity=0.80,line width= 0.4pt,line join=round,line cap=round,fill=fillColor,fill opacity=0.80] (309.76,117.88) circle (  0.89);

\path[draw=drawColor,draw opacity=0.80,line width= 0.4pt,line join=round,line cap=round,fill=fillColor,fill opacity=0.80] (309.76,120.69) circle (  0.89);

\path[draw=drawColor,draw opacity=0.80,line width= 0.4pt,line join=round,line cap=round,fill=fillColor,fill opacity=0.80] (309.76,122.36) circle (  0.89);

\path[draw=drawColor,draw opacity=0.80,line width= 0.4pt,line join=round,line cap=round,fill=fillColor,fill opacity=0.80] (309.76,114.56) circle (  0.89);

\path[draw=drawColor,draw opacity=0.80,line width= 0.4pt,line join=round,line cap=round,fill=fillColor,fill opacity=0.80] (309.76,120.00) circle (  0.89);

\path[draw=drawColor,draw opacity=0.80,line width= 0.4pt,line join=round,line cap=round,fill=fillColor,fill opacity=0.80] (309.76,143.49) circle (  0.89);
\definecolor{drawColor}{gray}{0.20}

\path[draw=drawColor,line width= 0.6pt,line join=round] (309.76,135.24) -- (309.76,140.41);

\path[draw=drawColor,line width= 0.6pt,line join=round] (309.76,130.29) -- (309.76,123.40);
\definecolor{fillColor}{RGB}{204,102,119}

\path[draw=drawColor,line width= 0.6pt,line join=round,line cap=round,fill=fillColor,fill opacity=0.80] (305.99,135.24) --
	(305.99,130.29) --
	(313.54,130.29) --
	(313.54,135.24) --
	(305.99,135.24) --
	cycle;

\path[draw=drawColor,line width= 1.1pt,line join=round] (305.99,133.11) -- (313.54,133.11);
\end{scope}
\begin{scope}
\path[clip] ( 58.21,174.26) rectangle (197.05,191.06);

\definecolor{drawColor}{gray}{0.10}

\node[text=drawColor,anchor=west,inner sep=0pt, outer sep=0pt, scale=  1] at 
(58.21,179.63) {(c) Data set \texttt{magic04}};
\end{scope}
\begin{scope}

\definecolor{drawColor}{gray}{0.10}

\node[text=drawColor,anchor=west,inner sep=0pt, outer sep=0pt, scale=  1] 
at 
(202.55,179.63) {(d) Data set \texttt{mushroom}};
\end{scope}
\begin{scope}
\path[clip] ( 58.21,266.78) rectangle (197.05,283.58);

\definecolor{drawColor}{gray}{0.10}

\node[text=drawColor,anchor=west,inner sep=0pt, outer sep=0pt, scale=  1] at 
(58.21,272.15) {(a) Data set \texttt{adult}};
\end{scope}
\begin{scope}

\definecolor{drawColor}{gray}{0.10}

\node[text=drawColor,anchor=west,inner sep=0pt, outer sep=0pt, scale=  1] at 
(202.55,272.15) {(b) Data set \texttt{drug consumption}};
\end{scope}
\begin{scope}
\path[clip] (  0.00,  0.00) rectangle (346.90,289.08);
\definecolor{drawColor}{RGB}{0,0,0}

\path[draw=drawColor,line width= 0.6pt,line join=round] ( 58.21,104.05) --
	(197.05,104.05);
\end{scope}
\begin{scope}
\path[clip] (  0.00,  0.00) rectangle (346.90,289.08);
\definecolor{drawColor}{gray}{0.20}

\path[draw=drawColor,line width= 0.6pt,line join=round] ( 71.65,101.30) --
	( 71.65,104.05);

\path[draw=drawColor,line width= 0.6pt,line join=round] ( 94.04,101.30) --
	( 94.04,104.05);

\path[draw=drawColor,line width= 0.6pt,line join=round] (116.43,101.30) --
	(116.43,104.05);

\path[draw=drawColor,line width= 0.6pt,line join=round] (138.83,101.30) --
	(138.83,104.05);

\path[draw=drawColor,line width= 0.6pt,line join=round] (161.22,101.30) --
	(161.22,104.05);

\path[draw=drawColor,line width= 0.6pt,line join=round] (183.62,101.30) --
	(183.62,104.05);
\end{scope}
\begin{scope}
\path[clip] (  0.00,  0.00) rectangle (346.90,289.08);
\definecolor{drawColor}{gray}{0.30}

\node[text=drawColor,rotate= 45.00,anchor=base east,inner sep=0pt, outer 
sep=0pt, scale=  0.88] at ( 75.93, 94.81) {\texttt{DIST-NUM}};

\node[text=drawColor,rotate= 45.00,anchor=base east,inner sep=0pt, outer 
sep=0pt, scale=  0.88] at ( 98.33, 94.81) {\texttt{DIST-CAT}};

\node[text=drawColor,rotate= 45.00,anchor=base east,inner sep=0pt, outer 
sep=0pt, scale=  0.88] at (120.72, 94.81) {\texttt{MISS-NUM}};

\node[text=drawColor,rotate= 45.00,anchor=base east,inner sep=0pt, outer 
sep=0pt, scale=  0.88] at (143.11, 94.81) {\texttt{OUTL-NUM}};

\node[text=drawColor,rotate= 45.00,anchor=base east,inner sep=0pt, outer 
sep=0pt, scale=  0.88] at (165.51, 94.81) {\texttt{CORR-CAT}};

\node[text=drawColor,rotate= 45.00,anchor=base east,inner sep=0pt, outer 
sep=0pt, scale=  0.88] at (187.90, 94.81) {\texttt{CORR-NUM}};
\end{scope}
\begin{scope}
\path[clip] (  0.00,  0.00) rectangle (346.90,289.08);
\definecolor{drawColor}{RGB}{0,0,0}

\path[draw=drawColor,line width= 0.6pt,line join=round] (202.55,104.05) --
	(341.40,104.05);
\end{scope}
\begin{scope}
\path[clip] (  0.00,  0.00) rectangle (346.90,289.08);
\definecolor{drawColor}{gray}{0.20}

\path[draw=drawColor,line width= 0.6pt,line join=round] (215.99,101.30) --
	(215.99,104.05);

\path[draw=drawColor,line width= 0.6pt,line join=round] (238.38,101.30) --
	(238.38,104.05);

\path[draw=drawColor,line width= 0.6pt,line join=round] (260.78,101.30) --
	(260.78,104.05);

\path[draw=drawColor,line width= 0.6pt,line join=round] (283.17,101.30) --
	(283.17,104.05);

\path[draw=drawColor,line width= 0.6pt,line join=round] (305.57,101.30) --
	(305.57,104.05);

\path[draw=drawColor,line width= 0.6pt,line join=round] (327.96,101.30) --
	(327.96,104.05);
\end{scope}
\begin{scope}
\path[clip] (  0.00,  0.00) rectangle (346.90,289.08);
\definecolor{drawColor}{gray}{0.30}

\node[text=drawColor,rotate= 45.00,anchor=base east,inner sep=0pt, outer 
sep=0pt, scale=  0.88] at (220.28, 94.81) {\texttt{DIST-NUM}};

\node[text=drawColor,rotate= 45.00,anchor=base east,inner sep=0pt, outer 
sep=0pt, scale=  0.88] at (242.67, 94.81) {\texttt{DIST-CAT}};

\node[text=drawColor,rotate= 45.00,anchor=base east,inner sep=0pt, outer 
sep=0pt, scale=  0.88] at (265.06, 94.81) {\texttt{MISS-NUM}};

\node[text=drawColor,rotate= 45.00,anchor=base east,inner sep=0pt, outer 
sep=0pt, scale=  0.88] at (287.46, 94.81) {\texttt{OUTL-NUM}};

\node[text=drawColor,rotate= 45.00,anchor=base east,inner sep=0pt, outer 
sep=0pt, scale=  0.88] at (309.85, 94.81) {\texttt{CORR-CAT}};

\node[text=drawColor,rotate= 45.00,anchor=base east,inner sep=0pt, outer 
sep=0pt, scale=  0.88] at (332.25, 94.81) {\texttt{CORR-NUM}};
\end{scope}
\begin{scope}
\path[clip] (  0.00,  0.00) rectangle (346.90,289.08);
\definecolor{drawColor}{RGB}{0,0,0}

\path[draw=drawColor,line width= 0.6pt,line join=round] ( 58.21,196.56) --
	( 58.21,266.78);
\end{scope}
\begin{scope}
\path[clip] (  0.00,  0.00) rectangle (346.90,289.08);
\definecolor{drawColor}{gray}{0.30}

\node[text=drawColor,anchor=base east,inner sep=0pt, outer sep=0pt, scale=  
0.88] at ( 53.26,213.66) {$10^{-2}$};

\node[text=drawColor,anchor=base east,inner sep=0pt, outer sep=0pt, scale=  
0.88] at ( 53.26,235.09) {$10^{+1}$};

\node[text=drawColor,anchor=base east,inner sep=0pt, outer sep=0pt, scale=  
0.88] at ( 53.26,256.52) {$10^{+4}$};
\end{scope}
\begin{scope}
\path[clip] (  0.00,  0.00) rectangle (346.90,289.08);
\definecolor{drawColor}{gray}{0.20}

\path[draw=drawColor,line width= 0.6pt,line join=round] ( 55.46,216.69) --
	( 58.21,216.69);

\path[draw=drawColor,line width= 0.6pt,line join=round] ( 55.46,238.12) --
	( 58.21,238.12);

\path[draw=drawColor,line width= 0.6pt,line join=round] ( 55.46,259.55) --
	( 58.21,259.55);
\end{scope}
\begin{scope}
\path[clip] (  0.00,  0.00) rectangle (346.90,289.08);
\definecolor{drawColor}{RGB}{0,0,0}

\path[draw=drawColor,line width= 0.6pt,line join=round] ( 58.21,104.05) --
	( 58.21,174.26);
\end{scope}
\begin{scope}
\path[clip] (  0.00,  0.00) rectangle (346.90,289.08);
\definecolor{drawColor}{gray}{0.30}

\node[text=drawColor,anchor=base east,inner sep=0pt, outer sep=0pt, scale=  
0.88] at ( 53.26,121.14) {$10^{-2}$};

\node[text=drawColor,anchor=base east,inner sep=0pt, outer sep=0pt, scale=  
0.88] at ( 53.26,142.57) {$10^{+1}$};

\node[text=drawColor,anchor=base east,inner sep=0pt, outer sep=0pt, scale=  
0.88] at ( 53.26,164.01) {$10^{+4}$};
\end{scope}
\begin{scope}
\path[clip] (  0.00,  0.00) rectangle (346.90,289.08);
\definecolor{drawColor}{gray}{0.20}

\path[draw=drawColor,line width= 0.6pt,line join=round] ( 55.46,124.17) --
	( 58.21,124.17);

\path[draw=drawColor,line width= 0.6pt,line join=round] ( 55.46,145.60) --
	( 58.21,145.60);

\path[draw=drawColor,line width= 0.6pt,line join=round] ( 55.46,167.04) --
	( 58.21,167.04);
\end{scope}
\begin{scope}
\path[clip] (  0.00,  0.00) rectangle (346.90,289.08);
\definecolor{drawColor}{RGB}{0,0,0}

\node[text=drawColor,anchor=base,inner sep=0pt, outer sep=0pt, scale=  1] at 
(199.80, 47.51) {EDA function};
\end{scope}
\begin{scope}
\path[clip] (  0.00,  0.00) rectangle (346.90,289.08);
\definecolor{drawColor}{RGB}{0,0,0}

\node[text=drawColor,rotate= 90.00,anchor=base,inner sep=0pt, outer sep=0pt, 
scale=  1] at ( 13.08,185.41) {};

\node[text=drawColor,rotate= 90.00,anchor=base,inner sep=0pt, outer sep=0pt, 
scale=  1] at ( 24.96,185.41) {relative error};

\node[text=drawColor,rotate= 90.00,anchor=base,inner sep=0pt, outer sep=0pt, 
scale=  1] at ( 36.84,185.41) {};
\end{scope}
\begin{scope}
\path[clip] (  0.00,  0.00) rectangle (346.90,289.08);
\definecolor{drawColor}{RGB}{0,0,0}

\node[text=drawColor,anchor=base west,inner sep=0pt, outer sep=0pt, scale=  1] 
at ( 65.03, 27) {mechanism};
\end{scope}
\begin{scope}
\path[clip] (  0.00,  0.00) rectangle (346.90,289.08);
\definecolor{drawColor}{gray}{0.20}

\path[draw=drawColor,line width= 0.6pt,line join=round,line cap=round] (148.31, 
22.81) --
	(148.31, 25.52);

\path[draw=drawColor,line width= 0.6pt,line join=round,line cap=round] (148.31, 
34.55) --
	(148.31, 37.26);
\definecolor{fillColor}{RGB}{68,119,170}

\path[draw=drawColor,line width= 0.6pt,line join=round,line 
cap=round,fill=fillColor,fill opacity=0.80] (141.54, 25.52) rectangle (155.09, 
34.55);

\path[draw=drawColor,line width= 0.6pt,line join=round,line cap=round] (141.54, 
30.03) --
	(155.09, 30.03);
\end{scope}
\begin{scope}
\path[clip] (  0.00,  0.00) rectangle (346.90,289.08);
\definecolor{drawColor}{gray}{0.20}

\path[draw=drawColor,line width= 0.6pt,line join=round,line cap=round] (249.41, 
22.81) --
	(249.41, 25.52);

\path[draw=drawColor,line width= 0.6pt,line join=round,line cap=round] (249.41, 
34.55) --
	(249.41, 37.26);
\definecolor{fillColor}{RGB}{204,102,119}

\path[draw=drawColor,line width= 0.6pt,line join=round,line 
cap=round,fill=fillColor,fill opacity=0.80] (242.63, 25.52) rectangle (256.19, 
34.55);

\path[draw=drawColor,line width= 0.6pt,line join=round,line cap=round] (242.63, 
30.03) --
	(256.19, 30.03);
\end{scope}
\begin{scope}
\path[clip] (  0.00,  0.00) rectangle (346.90,289.08);
\definecolor{drawColor}{RGB}{0,0,0}

\node[text=drawColor,anchor=base west,inner sep=0pt, outer sep=0pt, scale=  
1] at (168.69, 27.00) {interactive};
\end{scope}
\begin{scope}
\path[clip] (  0.00,  0.00) rectangle (346.90,289.08);
\definecolor{drawColor}{RGB}{0,0,0}

\node[text=drawColor,anchor=base west,inner sep=0pt, outer sep=0pt, scale=  
1] at (269.78, 27.00) {synthetic data};
\end{scope}
\end{tikzpicture}
 	\vspace{-3em}
	\caption{Accuracy of statistical functions given by the relative error.}
	\label{fig:mse}
	\vspace{-1ex}
\end{figure}
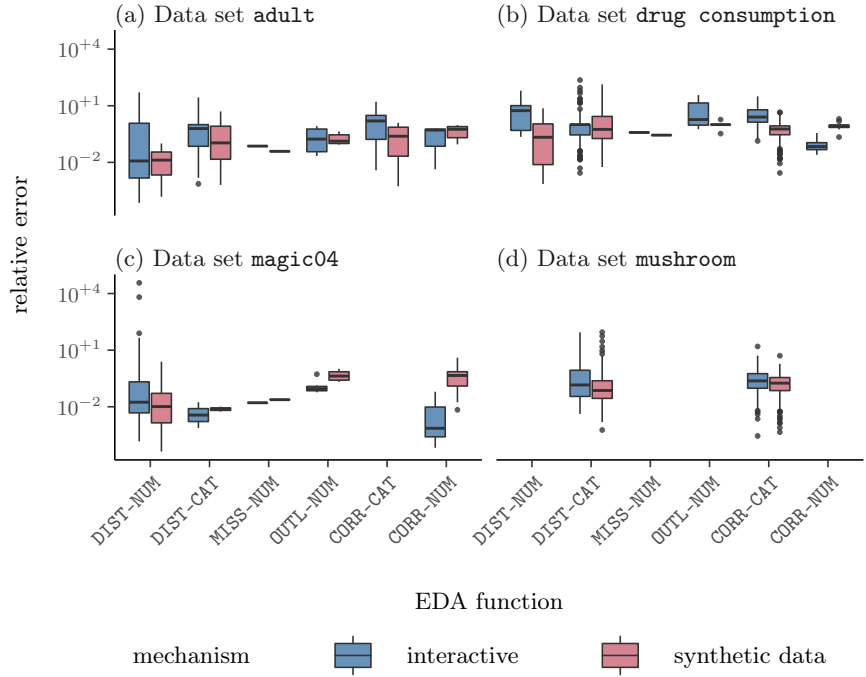

We measure the accuracy of our basic \ac{EDA} using the
relative error of each query.
\Cref{fig:mse} reports the relative error for each statistical function
visualized as
box plot containing the relative error for each query.
Overall, the synthetic data sets have a smaller relative error compared to the
interactive setting.
However, the relative error of the synthetic data sets for the numerical
correlation is higher compared to the interactive setting.
For the data set \texttt{magic04} (\Cref{fig:mse}c), we observe a similar 
effect for the
outliers. The high error occurs in the synthetic data sets due to the high
range of some numerical variables.
Remarkably, we obtain a high error for outliers for the interactive setting. 
This demonstrates that the relative error for the variables varies.

\subsection{Discussion}
\label{sec:discussion}
The results show that even basic investigations of an \ac{EDA} require a high
privacy budget in an interactive setting.
Therefore, an interactive analysis with both acceptable accuracy and an 
acceptable privacy guarantee is not possible.

Non-interactive mechanisms, such as differentially-private synthetic data sets,
can be used to increase accuracy and/or reduce the privacy budget.
Notably, the synthetic data set can be used directly to train a model without
dividing the privacy budget among \ac{EDA} and model generation.
However, the expressiveness of non-interactive mechanisms is limited to the
correlations used for generating the output. Therefore, these mechanisms are not
applicable for an interactive setting with random or unknown queries~\cite{DankarE13}.

In an \ac{EDA}, information is queried interactively, where one query
depends on the results of previous queries.
Grouping or limiting the queries to certain statistics,
a differentially-private \ac{EDA} might become feasible, though.
We therefore appeal to data analysts to agree on widely applicable
statistics that show the information necessary for model generation.
With our basic \ac{EDA}, we made a first attempt
to create such a collection of statistics.
A standardized set of statistical functions used in an \ac{EDA}
could be optimized to balance the privacy-utility trade-off.

\section{Conclusion}
\label{sec:conclusion}
In this paper we demonstrate the increase of the privacy loss and thus the
required budget for an
interactive differentially-private analysis. We argue that the \ac{EDA} should
be considered in privacy-preserving models, as it is an essential step in 
machine
learning.
In order to address the privacy-utility trade-off, we propose to agree on
standardized sets of \ac{EDA} functions and use the remaining privacy budget for
model creation.

\subsubsection*{Acknowledgement}
This work was supported by the Federal Ministry of Education and Research of 
Germany (project number 16KIS0909).

\bibliographystyle{splncs04}

\end{document}